%% file: main.tex
\documentclass[runningheads]{llncs}
\usepackage[T1]{fontenc}
\usepackage{graphicx}
\usepackage{amsmath}
\usepackage{amssymb}

\usepackage{subfigure}

\usepackage{booktabs}

\usepackage{hyperref}

\usepackage[textsize=small]{todonotes}

\begin{document}
%% Front %%
\input{front/title}
\maketitle              % typeset the header of the contribution
\input{front/abstract}

%% Sections %%
\input{sections/introduction}
\input{sections/related_works}
\input{sections/proposed_method}
\input{sections/experiments}

\input{sections/conclusion}
\input{sections/limitations}

%% Appendix %%
\appendix
\input{appendix/hyperparameters}
\bibliographystyle{bibliography/splncs04}
\bibliography{bibliography/ref}
\end{document}

%% file: front/title.tex
\title{Training-free Conditional Object Composition for Multi-object Image Generation with Diffusion Models}
\titlerunning{condComp}
\author{Hsuan-Fu Wang \and Hsuan-Tien Lin}
\authorrunning{H.-F. Wang and H.-T. Lin}
\institute{
    Graduate Institute of Networking and Multimedia,
    National Taiwan University\\
    \email{\{r11944074,htlin\}@csie.ntu.edu.tw}
}

%% file: front/abstract.tex
\begin{abstract}
Multi-object image generation in text-to-image diffusion models remains challenging, as existing approaches struggle to reliably generate all objects specified in a prompt. Compositional approaches that generate objects independently often fail to capture the interactions among them, whereas joint generation methods tend to suffer from inter-object interference.
% \htcomment{
% Compositional approaches that generate objects independently often fail to capture the interactions among them, whereas joint generation methods tend to suffer from inter-object interference.
% }
In this work, we propose a training-free framework, termed \textit{conditional object composition}, that relaxes joint generation methods by taking one step back toward compositional approaches.
% \htcomment{
% In this work, we propose a training-free framework, termed \textit{conditional object composition}, that relaxes joint generation methods by taking one step back toward compositional approaches.
% }
Our framework decomposes the generation of $N$ objects into two stages: it first generates a subset of $N-1$ objects and then composes the remaining object conditioned on the generated content.
% \htcomment{
% Our framework decomposes the generation of $N$ objects into two stages: it first generates a subset of $N-1$ objects and then composes the remaining object conditioned on the generated content.
% }
To enable effective composition, we adopt an inpainting-based formulation with layout guidance to preserve existing regions and encourage spatially coherent placement. To address the challenge of selecting which objects to generate separately, we introduce an attention-based scoring mechanism based on cross-attention maps and further explore the use of off-the-shelf object detectors. We evaluate our method on multi-object image generation using the widely adopted MS COCO dataset. Experimental results show that our approach achieves competitive or improved performance compared to existing methods, particularly as the number of objects increases. Further studies highlight the importance of the proposed design choices. The source code is available at \url{https://github.com/ShampooWang/condComp_imageGen/tree/main}

% while additional analysis suggests that our method can help alleviate challenges in several scenarios.
\end{abstract}

%% file: sections/introduction.tex
\section{Introduction}
\begin{figure*}[htb]
    \centering
    \subfigure[SDXL~\cite{podellsdxl} misses the required object "cat"]{
        \centering
        \includegraphics[width=0.47\linewidth]{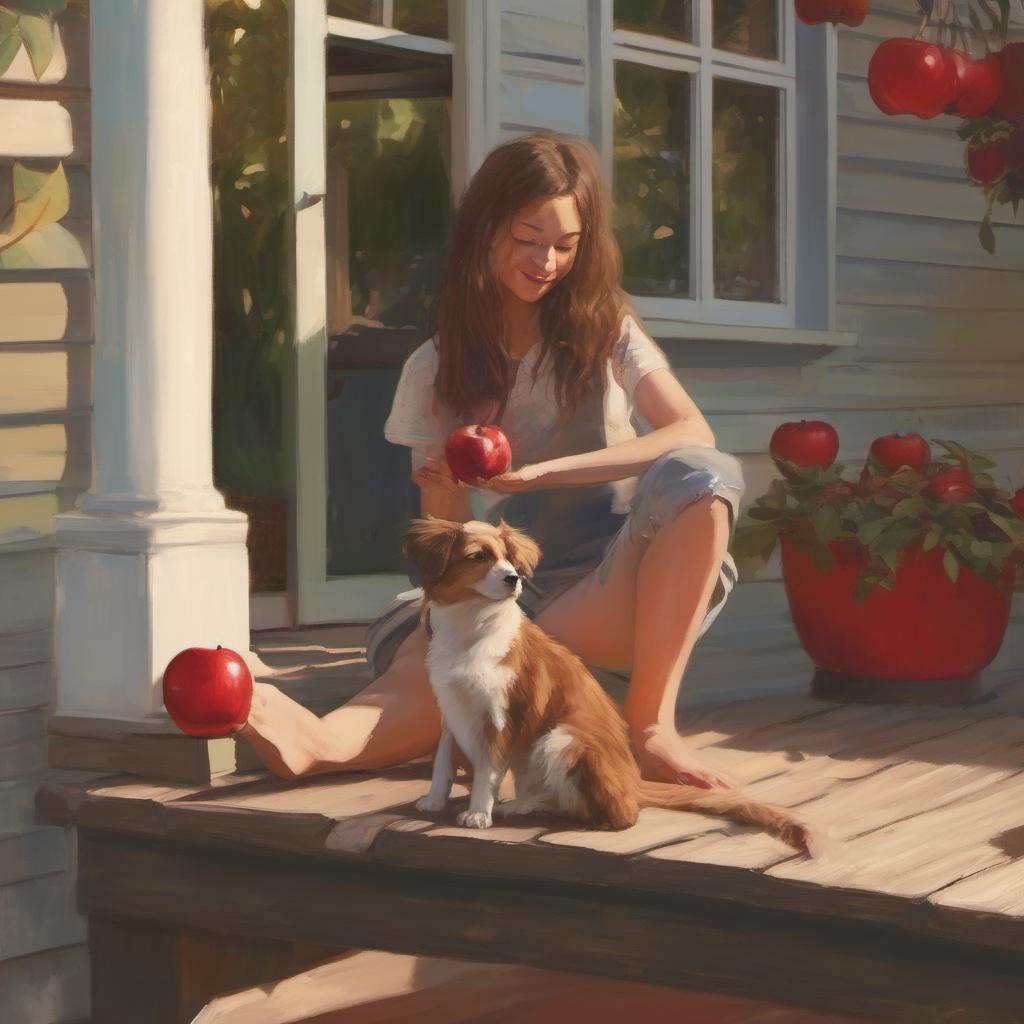}
        \label{fig:sdxl_failed_example}
    }
    \hfill
    \subfigure[Our method: adding the dog conditioned on generated objects (person, apple, and cat).]{
        \centering
        \includegraphics[width=0.47\linewidth]{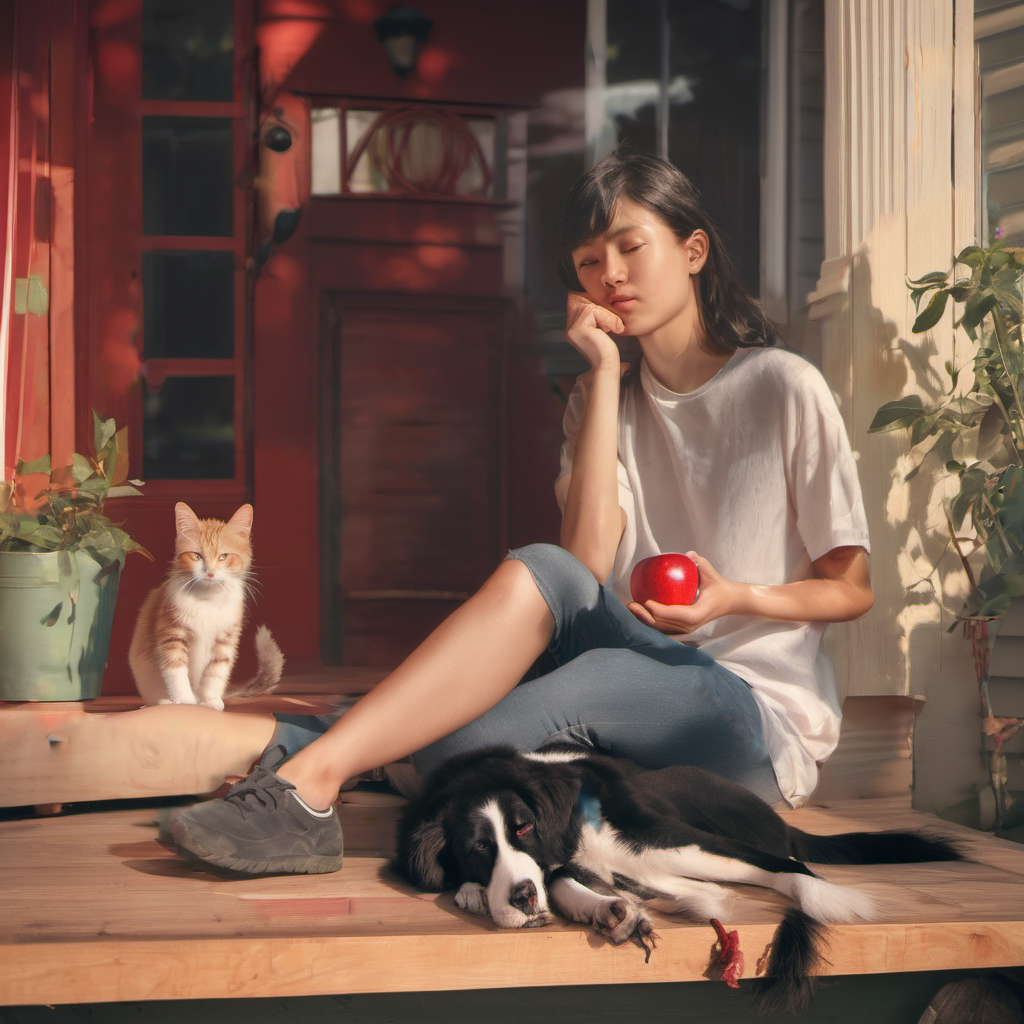}
        \label{fig:composition_example}
    }
    \caption{Comparison of multi-object image generation with required objects: person, apple, cat, and dog. 
    The base diffusion model fails to generate all specified objects, while our method improves completeness by composing objects conditioned on previously generated content. In this example, the model first generates a subset of objects (person, apple, and cat) and then adds the remaining object (dog) in a subsequent composition step.}
    % \label{fig:composition_example}
\end{figure*}

% \begin{figure}[t]
%     \centering
%     \begin{minipage}[t]{0.48\textwidth}
%         \includegraphics[width=\linewidth]{imgs/cat_dog_example.png}
%         (a) SDXL~\cite{podellsdxl} misses the required object ''cat.''
%     \end{minipage}
%     \hfill
%     \begin{minipage}[t]{0.48\textwidth}
%         \includegraphics[width=\linewidth]{imgs/condComp_expample2.png}
%         (b) Our method adds the dog conditioned on the generated person,
%         apple, and cat.
%     \end{minipage}
%     \caption{
%     Comparison of multi-object image generation for the required objects:
%     person, apple, cat, and dog.
%     (a) The base diffusion model fails to generate all specified objects.
%     (b) Our method first generates person, apple, and cat, and then composes
%     the remaining dog conditioned on the generated content.
%     }
%     \label{fig:composition_example}
% \end{figure}

\begin{figure*}[htb]
    \centering
    \subfigure[Success rate of multi-object image generation on MS COCO with varying numbers of objects.]{
        \centering
        \includegraphics[width=0.47\linewidth]{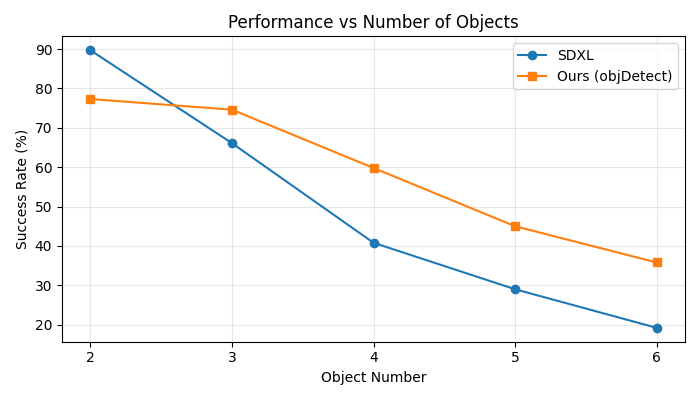}
        \label{fig:performance_vs_objects}
    }
    \hfill
    % Right: qualitative examples
    % \begin{minipage}[t]{0.48\textwidth}
    \subfigure[Failure cases of image editing with increasing editing strength.]{
        \centering
        \begin{minipage}[t]{0.15\linewidth}
            \includegraphics[width=\linewidth]{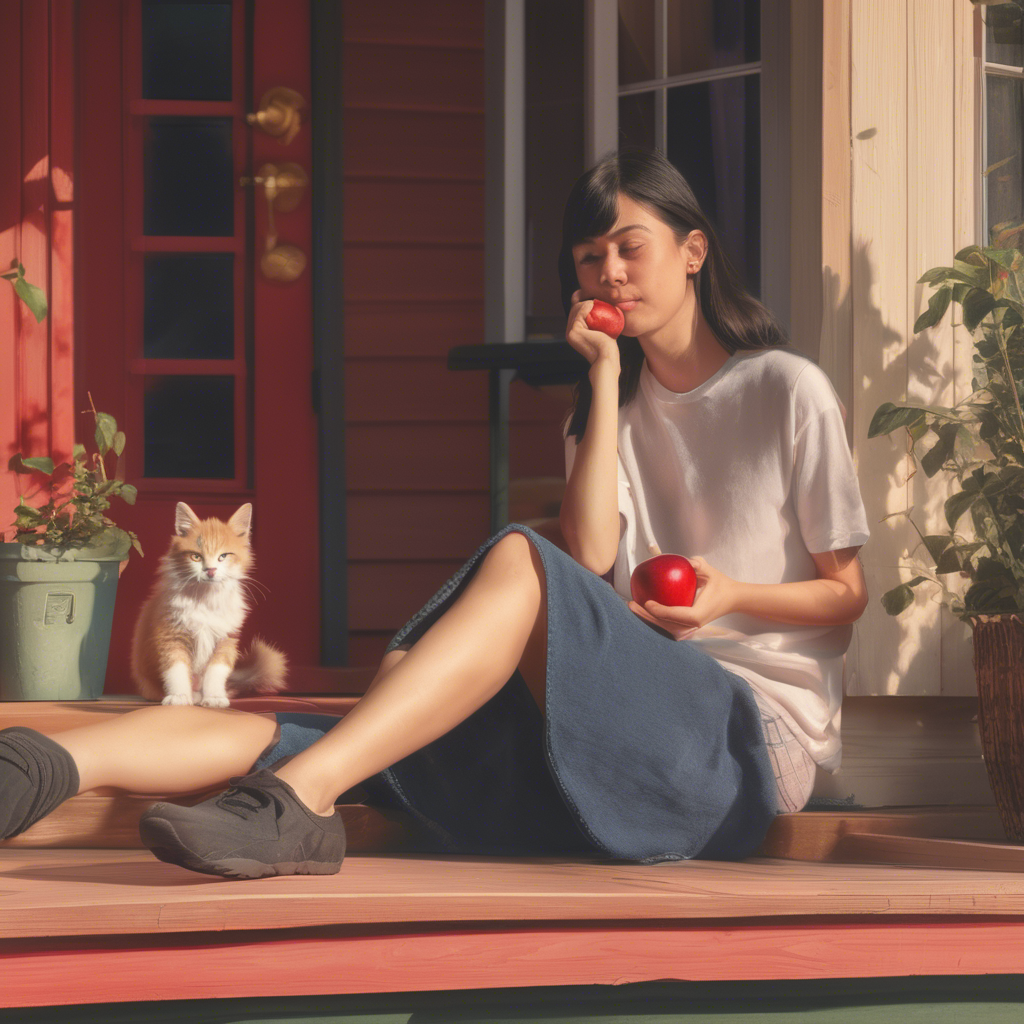}
        \end{minipage}
        \begin{minipage}[t]{0.15\linewidth}
            \includegraphics[width=\linewidth]{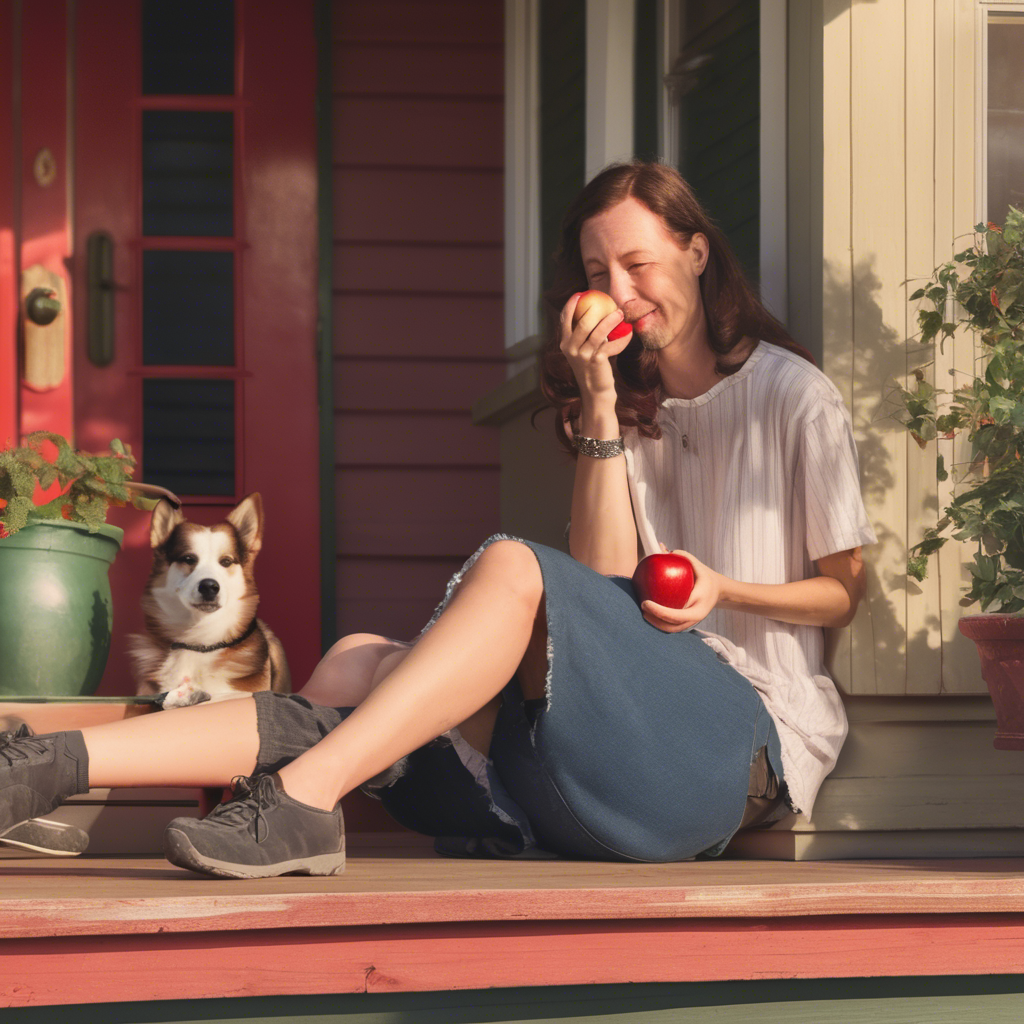}
        \end{minipage}
        \begin{minipage}[t]{0.15\linewidth}
            \includegraphics[width=\linewidth]{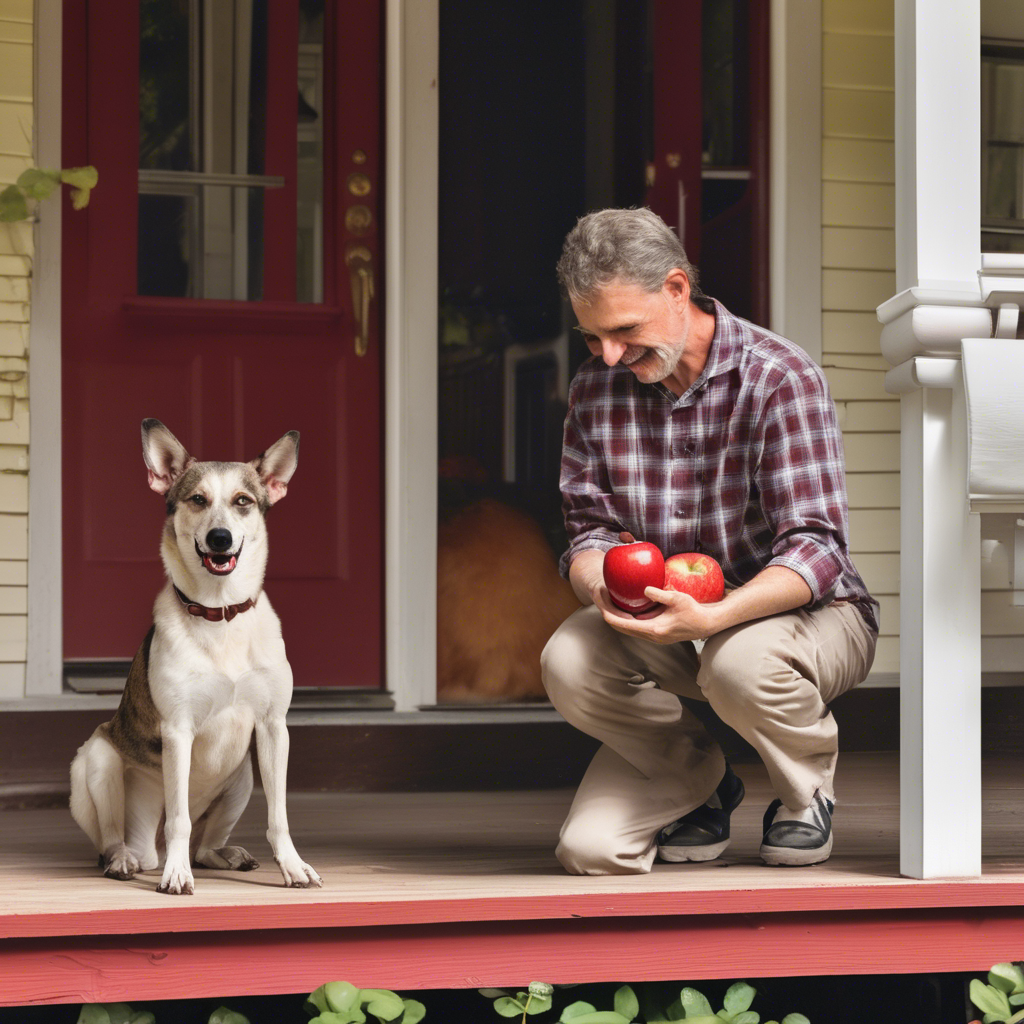}
        \end{minipage}
        % \caption{Failure cases of image editing with increasing editing strength.}
        \label{fig:editing_failure_cases}
    }
    % \caption{(a) Success rate of multi-object image generation on MS COCO with varying numbers of objects. Prompts are constructed from ground-truth object labels, and success requires all specified object categories to be present in the generated image. The base diffusion model exhibits a sharp performance degradation as the number of objects increases, whereas our method maintains higher success rates and shows improved robustness in complex multi-object scenarios. (b) Qualitative failure cases of general image editing for object composition. As the editing strength increases from left to right, the method transitions from under-editing, where the target object (dog) is not inserted, to over-editing, where previously generated content (cat) is increasingly altered.}
    \caption{
        (a) Success rate on MS COCO as the number of target objects increases. Our method consistently outperforms the base diffusion model, particularly in challenging multi-object scenarios. Success requires all specified object categories to be generated.
        (b) Failure cases of general image editing for object composition. Increasing the editing strength shifts the behavior from under-editing (missing the target object) to over-editing (modifying previously generated content).
    }    
    \label{fig:combined_results}
\end{figure*}

Recent advances in text-to-image diffusion models~\cite{podellsdxl,rombach2022sd,nichol2022glide,esser2024scaling,saharia2022imagen} have significantly improved the realism and diversity of generated images. However, generating images that contain all the objects specified in the text prompt remains a challenging problem~\cite{wang2024tokencompose}. 
% \htcomment{
%     maybe something more like "generating images that contain all the objects specified in the text prompt" (highlighting "all")
% }
In particular, when the number of objects increases, current models often fail to faithfully generate all required objects, as shown in Figure~\ref {fig:sdxl_failed_example}. This suggests that multi-object image generation poses challenges beyond those addressed by standard text-to-image generation. To better understand this limitation, we conduct a pilot study on the widely adopted multi-label dataset, MS COCO dataset~\cite{lin2014coco}, as shown in Figure~\ref{fig:performance_vs_objects}. 

Prompts are constructed using ground-truth object labels, and a generation is deemed successful only if all specified object categories appear in the image. By varying the number of objects in each prompt, we observe that the success rate drops substantially as the number of categories increases.

Multi-object image generation can be addressed by either training-based or training-free approaches. Training-based methods learn additional models or require fine-tuning on multi-label datasets, e.g., MS COCO~\cite{lin2014coco}, which can improve controllability but may require additional annotation processes~\cite{wang2024tokencompose} and incur extra training costs, and limit applicability to existing systems. In contrast, training-free approaches operate directly on pretrained diffusion models during inference, offering a more practical and flexible alternative. In this work, we focus on training-free approaches that operate on pretrained diffusion models without additional training.

Existing approaches~\cite{fengstructure,chen2024lg,chefer2023attend} primarily address this problem within a joint generation framework, where all objects are generated jointly in a single diffusion process by introducing refined cross-attention control or latent refinement techniques to improve object-level fidelity. For example, Structure Guidance~\cite{fengstructure} parses prompts using syntactic trees and applies additional cross-attention to the parsed components. Layout Guidance~\cite{chen2024lg} encourages cross-attention maps to focus within predefined bounding boxes through latent refinement. Attend-and-Excite~\cite{chefer2023attend} enhances object presence by increasing the minimum cross-attention saliency across objects, also via latent refinement.

However, despite these improvements, they remain fundamentally joint generation methods, which can lead to inter-object competition, where attention maps of different objects interfere with each other, making them increasingly difficult to reliably generate all objects as the number of objects grows. On the other hand, compositional methods~\cite{liu2022compositional} generate all objects individually to alleviate such competition. While this strategy circumvents disentanglement, it often fails to capture non-trivial interactions, e.g., spatial relationships between objects, leading to inconsistent or implausible compositions between objects.

To address these limitations, we adopt a strategy that relaxes joint generation methods by taking one step back toward compositional methods: instead of generating all $N$ objects jointly, we first generate $N-1$ objects and then compose the remaining object conditioned on the existing content. This formulation alleviates the difficulty of controlling multiple objects within a single generation process while allowing the model to leverage information from previously generated objects to capture interactions between objects. We refer to this framework as \textit{conditional object composition}. Such an approach transforms the original multi-object generation problem into two components: (1) composition, which integrates new objects into an existing scene, and (2) decomposition, which determines which objects should be generated separately rather than jointly.

However, this formulation introduces several challenges. First, object composition conditioned on previously generated content is non-trivial, as general image editting method~\cite{meng2022sdedit} tend to reuse or overwrite existing regions, as shown in Figure~\ref{fig:editing_failure_cases}, leading to object overlap or spatial conflicts. To address this, we perform composition in an inpainting-like manner~\cite{lugmayr2022repaint} by constructing a binary mask from cross-attention saliency, which preserves previously generated regions while encouraging new objects to occupy the remaining free space. Second, selecting an appropriate decomposition is challenging, as it requires identifying objects that are difficult to generate jointly. We observe that the salient regions of cross-attention maps capture spatial correspondences between text tokens and image regions, consistent with observations in prior works~\cite{tang2023daam,liu2024toward}. We leverage the spatial distribution of cross-attention values to guide the selection of objects for decomposition, and further explore the use of off-the-shelf object detectors as an alternative estimation mechanism.

% We evaluate the proposed method on multi-object image generation tasks using the widely adopted MS COCO dataset~\cite{lin2014coco}. Experimental results show that our approach achieves competitive or improved performance compared to existing methods. Further studies demonstrate that enforcing composition objectives, i.e., encouraging objects to occupy free space, is ineffective in joint generation with finer control through latent refinement. Our ablation studies further suggest that (1) effective decomposition requires careful selection, as naive strategies such as random selection can degrade performance, and (2) the joint generation strategy for the initial $N-1$ objects should consider not only the presence of these objects, but also the subsequent object composition stage. Additionally, our method shows advantages in handling spatially crowded and semantically ambiguous object configurations.

We evaluate the proposed method on multi-object image generation using the widely adopted MS COCO dataset~\cite{lin2014coco}. Experimental results demonstrate competitive or superior performance over existing methods. Ablation studies show that effective decomposition is essential, while simply enforcing free-space composition during joint generation is ineffective. They further highlight the importance of considering both joint generation and subsequent composition during decomposition. 
% Finally, our method effectively alleviates failures caused by spatial crowding and semantic ambiguity.

%% file: sections/related_works.tex
\section{Related Work}
\paragraph{\textbf{Finer control of text-to-image diffusion models.}}
Recent works have focused on achieving finer control over off-the-shelf text-to-image diffusion models~\cite{podellsdxl,rombach2022sd,nichol2022glide,esser2024scaling,saharia2022imagen} by refining cross-attention during inference. Methods such as Structure Guidance~\cite{fengstructure} parse input prompts using syntactic trees and apply additional cross-attention computations to the parsed components. Meanwhile, Layout Guidance~\cite{chen2024lg} and Attend-and-Excite~\cite{chefer2023attend} introduce latent refinement techniques to manipulate cross-attention maps. In particular, Layout Guidance~\cite{chen2024lg} enforces spatial constraints by encouraging attention maps to align with predefined bounding boxes, while Attend-and-Excite~\cite{chefer2023attend} enhances object presence by maximizing the minimum attention weight across target objects. Although these methods provide control at the object or token level, they operate within a joint generation framework, where cross-attention is computed and normalized over all objects simultaneously. As the number of objects increases, this shared attention mechanism introduces inter-object interference and competition for attention, which can limit the effectiveness of control. In contrast, our proposed method adopts a decomposed generation strategy, where objects are generated in different stages. By preserving previously generated regions (e.g., via masking), subsequent object generation is guided to occupy the remaining space, thereby reducing interference with existing content.

An alternative line of work decomposes multi-object generation into independent subproblems. Early works are based on energy-based models~\cite{du2021unsupervised,liu2021learning,nie2021controllable} and require additional training. Compositional Generation~\cite{liu2022compositional} extends this idea to text-to-image diffusion models by factorizing the generation process, modeling each object separately, and combining their conditional predictions via direct summation. While this strategy alleviates interference among objects, it often neglects their interactions, which may lead to inconsistent or implausible compositions when objects exhibit strong spatial or semantic dependencies~\cite{chefer2023attend}. In contrast, our formulation considers interactions by generating a subset of objects first and composing the remaining objects conditioned on them.

We note that some works~\cite{wang2024tokencompose,gaudicoind} achieve finer control by training on additional datasets. However, these approaches fall outside the scope of this work, as we specifically focus on training-free methods.

\paragraph{\textbf{Image Editing and Inpainting.}}
Image editing methods provide a complementary perspective for object composition by modifying or extending existing images while preserving their structure. Representative approaches for text-to-image diffusion models, such as SDEdit~\cite{meng2022sdedit}, introduce perturbations to an input image and perform the diffusion process conditioned on the editing target. Prompt-to-Prompt~\cite{hertz2023prompt} preserves image structure by reusing cross-attention maps during generation. However, general editing methods may struggle to precisely preserve object-level structure, as modifications can affect previously generated regions, as shown in Figure~\ref{fig:editing_failure_cases}.

To achieve more controlled preservation, we adopt an inpainting-based formulation~\cite{lugmayr2022repaint}, which uses spatial masks to specify regions to be preserved or modified during generation. This enables object composition while preserving previously generated content. However, we find that inpainting alone is insufficient, as newly added objects may be only partially generated or overlap with existing content. To better encourage objects to occupy available space, we incorporate the spatial control mechanism from layout guidance~\cite{chen2024lg} into the inpainting framework.

%% file: sections/proposed_method.tex
\section{Proposed method}
\subsection{Problem overview}
Given a text prompt that specifies a set of object categories $\mathcal{O} = \{o_1, o_2, \dots, o_N\}$, the goal is to generate an image that contains all objects in $\mathcal{O}$. Existing methods primarily adopt two strategies within a single-stage denoising process.~\cite{fengstructure,chen2024lg,chefer2023attend} perform joint generation with refined attention control, where all objects $\mathcal{O}$ are generated simultaneously based on a single conditional noise prediction. In contrast, compositional approaches~\cite{liu2022compositional} generate objects $o_1, o_2, \dots, o_N$ by computing object-specific noise predictions separately and combining them via direct summation without explicitly modeling interactions between all objects~\cite{chefer2023attend}.

In this work, we consider an alternative formulation that decomposes the generation process into two stages. Specifically, we first generate an image conditioned on a subset of objects $\mathcal{O'} \subset \mathcal{O}$, and then compose the remaining object $o_r \in \mathcal{O} \setminus \mathcal{O'}$ conditioned on the generated image. This reformulation alleviates the difficulty of jointly generating all objects while still enabling the model to capture interactions among them. It separates the problem into two components: (1) composition, which integrates new objects into an existing scene, and (2) decomposition, which determines which objects should be generated separately rather than jointly.

% This formulation introduces two key challenges. First, object composition requires integrating a new object into an existing image while preserving previously generated regions. Second, the choice of decomposition, i.e., selecting which object to generate separately, affects the overall generation quality. In the following sections, we describe our approach to addressing these challenges.

This formulation introduces two key challenges. First, object composition requires integrating a new object into an existing image while preserving previously generated regions. Second, the choice of decomposition, i.e., selecting which object to generate separately, affects the overall generation quality. To describe our method, let $A \in \mathbb{R}^{H \times W \times M}$ denote the aggregated cross-attention map, where $M$ is the number of text tokens, and let $A_{:,:,j} \in \mathbb{R}^{H \times W}$ denote the attention map associated with token $j$. Unless otherwise specified, $A$ is aggregated over the selected cross-attention layers and diffusion timesteps. We focus on attention maps corresponding to object tokens. We next describe our approach to composition and decomposition.

\subsection{Composition}
Given an image $x^\mathcal{O'}$ generated from a subset of objects $\mathcal{O'}$ in the first stage, our goal is to compose an additional object $o_r$ into the existing scene while preserving previously generated content in the second stage. To acheive this, we perform composition in an inpainting-like manner~\cite{lugmayr2022repaint}, where the generation of the new object $o_r$ is conditioned on a partially preserved image. 
% An overview of the proposed framework is illustrated in Figure~\ref{fig:composition_pipeline}.

\paragraph{\textbf{Mask Construction.}}
To preserve previously generated objects $\mathcal{O'}$ during composition, we construct a spatial mask $M \in \{0,1\}^{H \times W}$, where $M=0$ denotes preserved regions and $M=1$ denotes regions available for generating the new object $o_r$. We observe that cross-attention saliency provides a useful signal for spatial correspondences between text tokens and image regions, which is also suggested by prior works~\cite{liu2024toward,tang2023daam}. Motivated by this observation, we construct the mask to preserve generated objects based on cross-attention maps during the generation of $\mathcal{O'}$.

For each object $o_i \in \mathcal{O'}$, let $A_{:,:,j} \in \mathbb{R}^{H \times W}$ denote the cross-attention map corresponding to its associated token index $j$. We first apply min-max normalization to each map:
\begin{equation}
A'_{:,:,j} = \frac{A_{:,:,j} - \min(A_{:,:,j})}{\max(A_{:,:,j}) - \min(A_{:,:,j})}.
\label{eq:min_max_norm}
\end{equation}

We then aggregate the normalized maps across all objects:
\begin{equation}
A' = \frac{1}{|\mathcal{O'}|} \sum_{o_i \in \mathcal{O'}} A'_{:,:,j}.
\end{equation}

Finally, we obtain a binary mask $M$ by thresholding the aggregated attention map $A'$, where salient regions are assumed to correspond to object locations. Let $u \in \{1, 2, \dots,H\} \times \{1, 2, \dots,W\}$ denote a spatial location. The mask is defined as
\begin{equation}
M_u =
\begin{cases}
0, & \text{if } A'_u \ge \tau, \\
1, & \text{otherwise},
\end{cases}
\label{eq:mask_creation}
\end{equation}

where $\tau$ is a threshold corresponding to the top-$k$\% values of $A'$.

% \begin{figure*}[t]
%     \centering
%     \includegraphics[
%         width=\textwidth,     
%         trim={0 1cm 0 1.5cm},
%         clip
%     ]{imgs/decomp_v4.pdf}
%     \caption{
%     Illustration of the proposed conditional object composition framework 
%     by taking target object set $\mathcal{O} = \{\text{"person"}, 
%     \text{"apple"}, \text{"cat"}, \text{"dog"}\}$ as the example.
%     We evaluate all possible decompositions by selecting one object to be generated in the second stage. 
%     The left panel shows the attention-based strategy, where decomposition candidates are ranked according to joint generation and composition scores derived from cross-attention statistics. 
%     The right panel shows the object detector-based strategy, where decomposition quality is estimated using an external object detector applied to the composed result.
%     }
%     \label{fig:decomposition_pipeline}
% \end{figure*}

\paragraph{\textbf{Inpainting-based Composition with Layout Guidance.}}

Given the mask $M$, we compose the remaining object $o_r$ into the generated image using an inpainting-based formulation~\cite{lugmayr2022repaint} combined with Layout Guidance~\cite{chen2024lg}. The mask $M \in \{0,1\}^{H \times W}$ defines preserved and editable regions, where $M=0$ denotes preserved regions and $M=1$ denotes regions available for generating the new object $o_r$.

At each denoising timestep $t$ from $T$ to $0$, we first apply layout guidance to encourage the new object to occupy available space and avoid overlapping with existing content. Following~\cite{chen2024lg}, we define an energy function based on cross-attention maps that penalizes placing the target object in regions already occupied by existing objects. Specifically, given the cross-attention map $A$, the mask $M$, and the token indices $j'$ corresponding to the object $o_r$, the energy function is defined as
\begin{equation}
    E(A, M, j') = \left( 1 - \frac{\sum_{u} A_{u j'} \, M_u}{\sum_{u} A_{u j'}} \right)^2.
\end{equation}

The latent is then updated via gradient descent on this energy:
\begin{equation}
z_t \leftarrow z_t - \eta \nabla_{z_t} E(A, M, j'),
\label{eq:free_space_guidance}
\end{equation}
where $\eta$ controls the step size.

After the latent update, we perform inpainting to integrate the new object $o_r$ while preserving existing content. The preserved region (defined by $M=0$) is diffused from the latent of the previously generated image $x^{\mathcal{O}'}$, denoted by $z^{\mathcal{O}'}$, while the editable region (defined by $M=1$) is updated using the diffusion model. The resulting latent at timestep $t$ is obtained by combining the two regions:
\begin{equation}
z_t = M \odot z_t + (1 - M) \odot z^{\mathcal{O}'}_t,
\label{eq:inpaint_update}
\end{equation}
where $\odot$ denotes element-wise multiplication.

% This process preserves previously generated objects while allowing the new object to be synthesized in the remaining regions. We then proceed with the standard denoising update in Eq.~\eqref{eq:diffusion_update}, and repeat Eqs.~\eqref{eq:free_space_guidance} and~\eqref{eq:inpaint_update} at each timestep $t$ from $T$ to $0$.

This process preserves previously generated objects while allowing the new object to be synthesized in the remaining regions. We then proceed with the standard denoising update, and repeat Eqs.~\eqref{eq:free_space_guidance} and~\eqref{eq:inpaint_update} at each timestep $t$ from $T$ to $0$.

\subsection{Decomposition}
A key component of our framework is to determine which object should be generated separately. Intuitively, objects that are difficult to generate jointly often exhibit strong interference with other objects, which can be reflected in their cross-attention distributions. 
% An overview of the proposed framework is illustrated in Figure~\ref{fig:decomposition_pipeline}.

To quantify this effect, we define an attention-based scoring function that measures the degree of interference between objects. Let $A_{:,:,i} \in \mathbb{R}^{H \times W}$ denote the cross-attention map corresponding to object $o_i$. We use the cross-attention maps at diffusion timesteps $t_{\text{joint}}$ and $t_{\text{comp}}$ for the joint generation and composition stages, respectively. The attention map is then normalized using min-max normalization to obtain $A'_{:,:,i}$ in Eq.~\ref{eq:min_max_norm}. To suppress noisy responses, we further apply a threshold $\tau'$ and set normalized attention values below $\tau'$ to zero. We denote the resulting attention value at spatial location $u$ as $A'_{ui}$.

For a given subset $\mathcal{O'}$, we compute a score for each object $o_i \in \mathcal{O'}$:
\begin{equation}
s_i = \frac{\sum_{u} \left(A'_{ui} - \frac{1}{|\mathcal{O'}| - 1} \sum_{o_j \in \mathcal{O'},\, j \neq i} A'_{uj}\right)}{\sum_{u} A'_{ui}}.
\end{equation}

This score $s_i$ measures how well the attention of object $o_i$ is separated from that of other objects. A lower score indicates that the attention of $o_i$ overlaps significantly with other objects, suggesting that it is more difficult to generate jointly.

To obtain a global measure of the joint generation difficulty of the subset $\mathcal{O'}$, we take the minimum score across all objects:
\begin{equation}
s_{\text{joint}} = \min_{i} s_i,
\label{eq:joint_score}
\end{equation}

After generating the subset $\mathcal{O'}$, we further evaluate whether the newly added object $o_r$ is placed in appropriate regions. Let $A_{ur}$ denote the attention map of the added object. We define a composition score as
\begin{equation}
s_{\text{comp}} = \frac{\sum_{u} A_{ur} \, M(u)}{\sum_{u} A_{ur}},
\label{eq:comp_score}
\end{equation}
which measures the proportion of attention mass located in the editable region defined by the mask $M$.

Finally, for each candidate object $o_r \in \mathcal{O}$, we define an overall score as
\begin{equation}
s_{o_r} = s_{\text{joint}} \cdot s_{\text{comp}}.
\label{eq:attn_score}
\end{equation}

We select the object $o_r$ with the highest score $s_{o_r}$ for decomposition. While our main formulation relies on attention-based estimates, the decomposition strategy is not restricted to this choice. One may also incorporate external object detection models~\cite{khanam2024yolov11,li2021benchmarking,carion2020detr,zhao2024rtdetr} to guide the selection of $o_r$, for example, by estimating object presence or localization confidence. We explore such a variant in the experiments.

%% file: sections/experiments.tex
\section{Experiments}
\subsection{Experimental Setup}
\paragraph{\textbf{Evaluation Setup.}}
We evaluate our method on multi-object image generation using the SDXL base model~\cite{podellsdxl}. Prompts are constructed from ground-truth object categories in the MS COCO dataset~\cite{lin2014coco}. Specifically, we sample object sets $\mathcal{O} = \{o_1, o_2, \dots, o_N\}$ from COCO images, ensuring that all objects in $\mathcal{O}$ co-occur in the same image, rather than forming arbitrary combinations of categories. For each set $\mathcal{O}$, we construct prompts of the form: \emph{``A photo with realistic lighting, showing a $o_1$, $o_2$, $\dots$, $o_N$''}. Evaluation is conducted across varying numbers of objects per prompt to study scalability. For each setting, we generate 100 samples per seed across 5 random seeds. All methods are evaluated under the same sampling configuration for a fair comparison. Additional configuration details are provided in the Appendix~\ref{apd:first}.

\paragraph{\textbf{Evaluation Metric.}}
We measure performance using \emph{success rate}, defined as the percentage of generated images that contain all specified object categories in the input prompt. To automatically determine object presence, we employ a pre-trained object detector, YOLOv11-L~\cite{khanam2024yolov11}, to detect objects in the generated images. A generation is considered successful if all target object categories are detected.

\paragraph{\textbf{Baselines.}}
We compare our method against representative approaches from two categories. 
(1) \emph{Joint generation methods with finer control}, including Structure Guidance~\cite{fengstructure}, Layout Guidance~\cite{chen2024lg}, and Attend-and-Excite~\cite{chefer2023attend}, which generate all objects with refined attention control. 
(2) \emph{Compositional methods}, such as Compositional Generation ~\cite{liu2022compositional}, which decompose each object independently and compose them without spatial control. All baselines are implemented using the same underlying diffusion model, SDXL~\cite{podellsdxl}.

\begin{table}[t]
\centering
\caption{Success rate (\%) of multi-object image generation on MS COCO with varying number $N$ of objects. Results are reported as mean $\pm$ standard deviation over 5 seeds.}
\label{tab:coco_results}
\small
\begin{tabular}{lccc}
\toprule
Method & $N = 4$ & $N = 5$ & $N = 6$ \\
\midrule
SDXL & 40.8 $\pm$ 2.8 & 29.0 $\pm$ 3.0 & 19.2 $\pm$ 3.3 \\
Compositional Generation & 1.2 $\pm$ 1.2 & 0.4 $\pm$ 0.5 & 0.0 \\
Structure Guidance & 33.0 $\pm$ 0.9 & 22.8 $\pm$ 3.0 & 15.2 $\pm$ 1.7 \\
Layout Guidance & 42.6 $\pm$ 3.9 & 26.8 $\pm$ 6.0 & 19.0 $\pm$ 2.5 \\
Attend-and-Excite & 42.6 $\pm$ 3.7 & 29.0 $\pm$ 5.3 & 21.2 $\pm$ 2.7 \\
\midrule
condComp (attnScore) & \underline{48.4 $\pm$ 4.3} & \underline{34.6 $\pm$ 4.4} & \underline{25.2 $\pm$ 4.7} \\
condComp (objDetect) & \textbf{59.8 $\pm$ 3.2} & \textbf{45.0 $\pm$ 6.0} & \textbf{35.8 $\pm$ 5.3} \\
\bottomrule
\end{tabular}
\end{table}

\begin{figure}[t]
\centering
\includegraphics[width=0.9\linewidth]{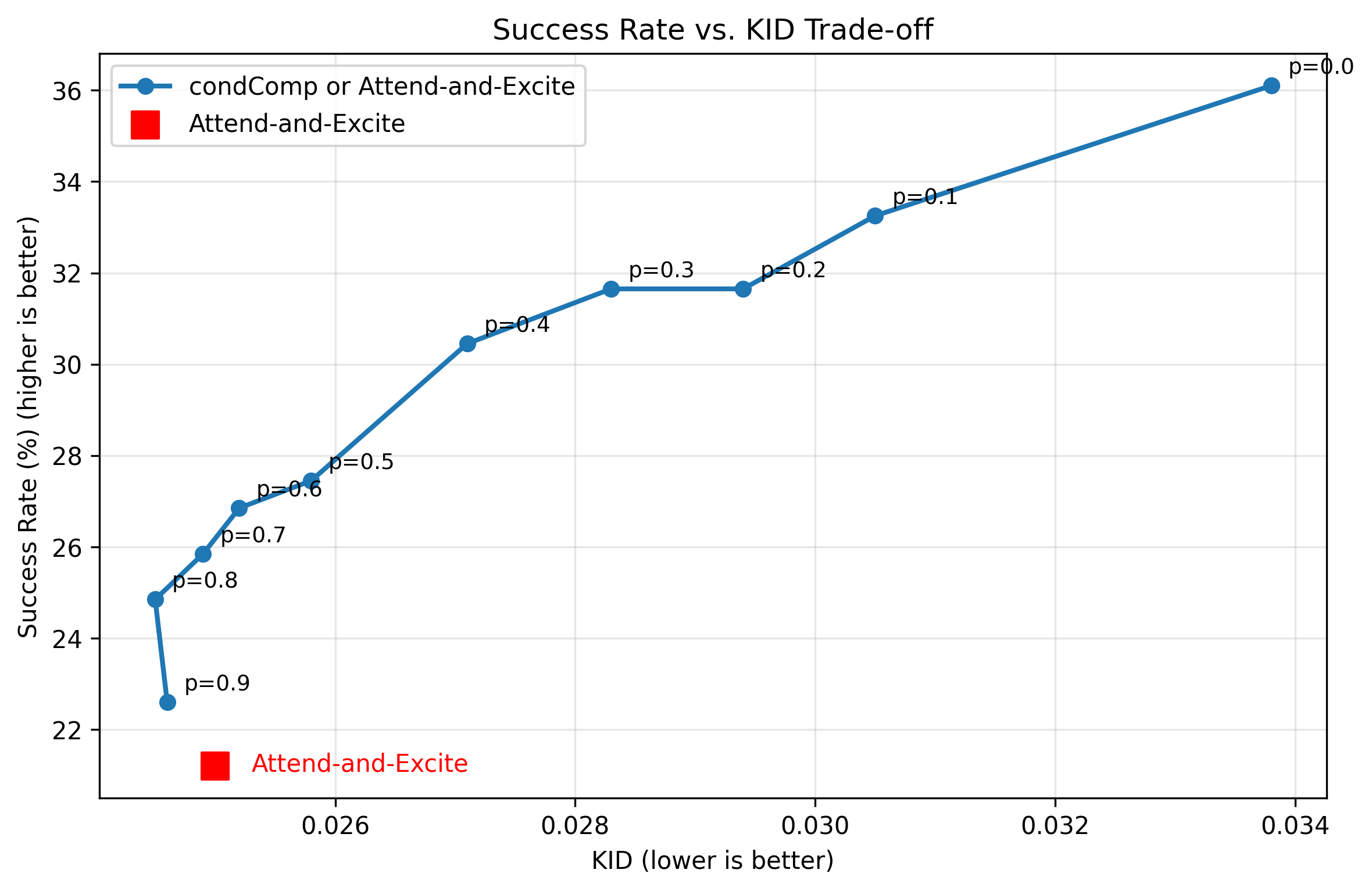}
\caption{Trade-off between success rate and realism measured by KID. Higher success rate and lower KID correspond to better performance. We interpolate between Attend-and-Excite~\cite{chefer2023attend} and our method by randomly switching to it with probability $p$. Lower $p$ corresponds to using conditional composition more frequently.}
\label{fig:kid_tradeoff}
\end{figure}

\subsection{Results on Multi-Object Generation}
\paragraph{\textbf{Comparison with Existing Methods.}}Table~\ref{tab:coco_results} reports the success rates of different methods on multi-object generation using MS COCO across varying numbers of objects. As the number of objects increases, all methods exhibit a noticeable decline in performance, reflecting the increasing difficulty of generating multiple objects. Among the baselines, Compositional Generation performs substantially worse than other baselines across all settings, highlighting the limitations of independently combining object-specific predictions without modeling interactions, particularly in more challenging scenarios with larger numbers of objects and complex real-world categories. We note that prior Compositional Generation was primarily evaluated on simpler settings, such as two to three objects or toy object categories (e.g., spheres and cubes).

We further compare two variants of our method: \textit{condComp (attnScore)}, which relies on attention-based estimation (Eq.~\ref{eq:attn_score}) for decomposition, and \textit{condComp (objDetect)}, which uses an external object detector. For the latter, we adopt Faster R-CNN~\cite{li2021benchmarking} to simulate a moderately accurate detector in practical scenarios. The attnScore variant achieves competitive performance when the number of objects is medium (e.g., 4 objects), and demonstrates clearer improvements as the number of objects increases. This trend suggests that decomposing the generation process and performing composition in a staged manner can be beneficial in more challenging multi-object settings. The objDetect variant consistently achieves higher success rates, indicating that a more accurate estimation for decomposition can further improve performance. 

Overall, these results show that our approach provides a competitive or improved alternative to existing methods, particularly in settings with a larger number of objects where joint generation becomes more challenging.

\paragraph{\textbf{Trade-off Between Success Rate and Realism.}}\label{para:kid_tradeoff}
We further analyze the trade-off between object completeness and image realism. Since the number of reference multi-object images is limited, we adopt Kernel Inception Distance (KID)~\cite{binkowski2018kid}, which is more suitable for small-sample evaluation than widely adopted FID~\cite{heusel2017fid}. Figure~\ref{fig:kid_tradeoff} reports the success rate and KID under different probabilities of switching to joint generation using Attend-and-Excite~\cite{chefer2023attend}. Specifically, at each generation, we perform conditional composition (objDetect) with probability $1-p$ and joint generation with probability $p$.
As the probability of conditional composition increases (smaller $p$), the success rate consistently improves, while the KID score gradually degrades. Notably, in several intermediate settings, such a hybrid strategy achieves substantially higher success rates while maintaining KID scores comparable to Attend-and-Excite. This suggests that conditional composition can improve object completeness without severely sacrificing image realism in some cases.

The transition between different probabilities is relatively smooth, indicating that hybrid generation provides a controllable trade-off between realism and object completeness. These results further suggest that conditional composition and joint generation provide complementary advantages for multi-object generation.

\begin{table}[t]
\centering
\caption{
Effect of applying the composition objective within joint generation frameworks.
Results are reported as success rate (\%) (mean $\pm$ standard deviation over 5 seeds).
}
\label{tab:joint_composition}

\small
\begin{tabular}{lccc}
\toprule
Method & $N = 4$ & $N = 5$ & $N = 6$ \\
\midrule
SDXL 
& 40.8 $\pm$ 2.8 
& 29.8 $\pm$ 3.0
& 19.2 $\pm$ 3.3 \\

SDXL + composition 
& 40.8 $\pm$ 2.9
& 28.0 $\pm$ 3.6 
& 18.6 $\pm$ 3.4 \\
\midrule
Attend-and-Excite 
& \textbf{42.6 $\pm$ 3.7} 
& \textbf{29.0 $\pm$ 5.3} 
& \textbf{21.2 $\pm$ 2.7} \\

Attend-and-Excite + composition 
& 40.6 $\pm$ 6.6
& 26.8 $\pm$ 6.1 
& 19.0 $\pm$ 2.5 \\
\bottomrule
\end{tabular}
\end{table}

\begin{table}[t]
\centering
\caption{
Effect of decomposition strategy.
Results are reported as success rate (\%) (mean $\pm$ standard deviation over 5 seeds).
}
\label{tab:decomposition_ablation}

\small
\begin{tabular}{lccc}
\toprule
Method & $N = 4$ & $N = 5$ & $N = 6$ \\
\midrule
condComp (random)
& 34.8 $\pm$ 2.9 
& 27.2 $\pm$ 3.6 
& 18.0 $\pm$ 7.1 \\

condComp (attnScore)
& \textbf{48.4 $\pm$ 4.3} 
& \textbf{34.6 $\pm$ 4.4} 
& \textbf{25.2 $\pm$ 4.7} \\
\bottomrule
\end{tabular}
\end{table}

\subsection{Composition Objective in Joint Generation}
As joint generation methods with finer control, such as Attend-and-Excite~\cite{chefer2023attend}, are strong baselines, we first investigate whether the composition objective—i.e., placing an object in free space suggested by other $N-1$ objects—can be achieved within a joint generation framework with similar mechanisms. Specifically, at each timestep of joint generation, we apply the same layout guidance objective (Eq.~\ref{eq:free_space_guidance}) to the object with the lowest composition score (Eq.~\ref{eq:comp_score}). The mask $M$ is constructed using the attention maps of the other objects. We incorporate this objective into baseline models, including the SDXL~\cite{podellsdxl} and Attend-and-Excite. As shown in Table~\ref{tab:joint_composition}, adding the composition objective consistently degrades performance across different numbers of objects. This suggests that directly enforcing spatial placement constraints within a joint generation process is challenging. One possible reason is that the notion of free space is not well-defined during joint generation, as attention maps of different objects may interfere with each other and evolve throughout the denoising process.

\subsection{Ablation Studies}
\paragraph{\textbf{Effect of Decomposition Strategy.}}
We next examine the importance of selecting an appropriate object for decomposition. To this end, we compare our method with a variant, denoted as \textit{condComp (random)} in Table~\ref{tab:decomposition_ablation}, which randomly selects the object to be generated in the second stage. The results show that random decomposition leads to noticeably worse performance, in some cases even below the base model. This indicates that the choice of decomposition plays a critical role, and that not all objects are equally suitable to be generated separately. In contrast, our attention-based scoring method provides a more reliable way to identify objects that are difficult to generate jointly, leading to improved performance.

\paragraph{\textbf{Effect of Joint Generation Strategy.}}
We further investigate whether replacing the joint generation stage with Attend-and-Excite~\cite{chefer2023attend} can be improved. Table~\ref{tab:joint_ae_stagewise} compares using the base diffusion model and Attend-and-Excite for the initial joint generation stage of $N-1$ objects. Incorporating Attend-and-Excite improves the KID score, while slightly reducing the overall success rate. This observation is consistent with the trade-off between realism and object completeness discussed in paragraph~\ref{para:kid_tradeoff}, suggesting that stronger joint generation method can improve image quality but may interfere with subsequent composition

To better understand where failures occur in the proposed framework, we separately analyze the success rates of the joint generation and composition stages in Table~\ref{tab:joint_ae_stagewise}. Replacing the joint generation stage with Attend-and-Excite leads to only a minor reduction in the joint generation success rate ($-0.89\%$), while the degradation is more pronounced in the composition stage ($\textbf{-3.51\%}$). This suggests that, in our proposed method, jointly generating the initial $N-1$ objects should also account for the subsequent object-composition stage.

\begin{table}[t]
\centering
\caption{
Effect of using Attend-and-Excite in the joint generation stage for $N=6$. 
Results are reported as mean $\pm$ standard deviation over 5 seeds.
}
\label{tab:joint_ae_stagewise}

\small
\begin{tabular}{lcccc}
\toprule
& \multicolumn{2}{c}{Overall} & \multicolumn{2}{c}{Stage-wise} \\
\cmidrule(lr){2-3}
\cmidrule(lr){4-5}
Method & Success Rate & KID & Joint $(N-1)$ & Composition \\
\midrule
Attend-and-Excite 
& 21.2 $\pm$ 2.7 
& \textbf{0.025 $\pm$ 0.004}
& -- & -- \\

condComp (SDXL) 
& \textbf{35.8 $\pm$ 5.3}
& 0.034 $\pm$ 0.006
& \textbf{46.3 $\pm$ 4.2}
& \textbf{57.5 $\pm$ 7.3} \\

condComp (Attend-and-Excite)
& \underline{32.8 $\pm$ 2.2}
& \underline{0.032 $\pm$ 0.005}
& 45.4 $\pm$ 3.4
& 54.0 $\pm$ 2.6 \\
\bottomrule
\end{tabular}
\end{table}

%% file: sections/conclusion.tex
\section{Conclusion}

In this work, we study multi-object image generation in text-to-image diffusion models from a training-free perspective. Existing compositional approaches that generate objects independently often fail to capture interactions among objects, while joint generation methods tend to suffer from inter-object interference. To address these limitations, we propose \textit{conditional object composition}, a framework that relaxes joint generation by taking one step toward compositional generation. Specifically, the proposed framework decomposes the generation of $N$ objects into two stages: it first jointly generates $N-1$ objects and then composes the remaining object conditioned on the generated content. To enable effective composition, we adopt an inpainting-based formulation with layout guidance to preserve existing regions and encourage spatially coherent placement of newly added objects. To address the challenge of selecting suitable objects for decomposition, we further introduce an attention-based scoring mechanism based on cross-attention maps and explore the use of off-the-shelf object detectors. Experiments on MS COCO demonstrate that our method achieves competitive or improved performance compared to existing methods, particularly as the number of objects increases. Further studies highlight the importance of the proposed design choices. 

% while additional analysis suggests that our framework can help alleviate challenges in spatially crowded and semantically ambiguous scenarios.

%% file: sections/limitations.tex
\section{Limitations}
Our method introduces additional computational overhead due to its sequential formulation. In practice, it requires approximately 3$\times$–6$\times$ longer inference time than single-stage methods such as Attend-and-Excite~\cite{chefer2023attend} in our experiments. In addition, the quality of composition is not always ideal: newly added objects may appear partially generated, and their placement can be spatially unnatural in some cases. Future work may explore more efficient generation strategies and improved composition mechanisms to produce more coherent and natural results.

%% file: appendix/hyperparameters.tex
\section{Appendix}
\paragraph{\textbf{Hyper-parameters in Experiments}}\label{apd:first}
We use $T=50$ diffusion timesteps for all methods. For the proposed method, each generation stage also uses $T=50$ timesteps. The classifier-free guidance scale is set to the default value of 5 for all methods.

For methods based on latent refinement, including Layout Guidance~\cite{chen2024lg}, Attend-and-Excite~\cite{chefer2023attend}, and the proposed method, we optimize cross-attention layers during the denoising process. Following Layout Guidance~\cite{chen2024lg}, we optimize the cross-attention layers in the middle and up blocks for Layout Guidance and the proposed method. In Eq.~\ref{eq:free_space_guidance}, we set the guidance scale $\eta$ to 50 for the proposed method, while for Layout Guidance and Attend-and-Excite we tuned the refinement scale factor to $\eta=30$. All experiments are conducted on a machine equipped with four NVIDIA V100 GPUs, each with 32GB of memory. We found that optimizing both the middle and up blocks for Attend-and-Excite frequently resulted in out-of-memory errors under this setting. Therefore, for Attend-and-Excite, we optimize only cross-attention layers 70--81, which yielded the best performance in our preliminary experiments.

For decomposition, we use the cross-attention maps at diffusion timestep
$t_{\text{joint}} = 20$ for the joint generation stage and
$t_{\text{comp}} = 50 = T$ for the composition stage. After min-max normalization,
we apply a threshold $\tau' = 0.6$ and set attention values below the threshold
to zero.